\documentclass[a4paper,11pt,twocolumn,twoside]{article}
\usepackage{graphicx}
\usepackage{sepln_en}
\usepackage{fullname}
\usepackage[utf8]{inputenc}
\usepackage{multirow}
\usepackage{booktabs}
\usepackage{graphicx}
\usepackage{url}
\usepackage{lipsum}
\usepackage{textgreek}

\input epsf

\title{LyricSIM: A novel dataset and benchmark for similarity detection in Spanish song lyrics\thanks{Accepted to Congreso Internacional de la Sociedad Española para el Procesamiento del Lenguaje Natural 2023 (SEPLN2023).}}

\author {\textbf{Alejandro Benito-Santos$^1$}, \textbf{Adrián Ghajari$^2$}, \textbf{Pedro Hernández$^3$}, \textbf{Víctor Fresno$^2$}\\ \textbf{Salvador Ros$^2$}, \textbf{Elena González-Blanco$^3$}\\
 $^1$Universidad de Salamanca\\
 $^2$Universidad Nacional de Educacion a Distancia\\
 $^3$IE University\\
 abenito@usal.es \\}

\seplntranstitle{LyricSIM: Un nuevo dataset y benchmark para la detección de similitud en letras de canciones en español}

\seplnclave{similitud semántica en textos, tarea de anotación, conjunto de datos, letras de canciones}

\seplnresumen{En este trabajo presentamos un nuevo conjunto de datos y \textit{benchmark} orientados a la tarea de similitud semántica en letras de canciones. Nuestro conjunto de datos, originalmente formado por 2775 pares de canciones en Español, fue anotado en un experimento de anotación colectivo por 63 anotadores nativos. Después de recoger y refinar los datos para asegurar un alto grado de consenso e integridad en los datos, obtuvimos 676 pares anotados de alta calidad que fueron empleados para evaluar el rendimiento de diferentes modelos del lenguaje monolingües y multilingües pertenecientes al estado del arte. En consecuencia, obtuvimos unos resultados base que esperamos sean de utilidad a la comunidad en todas aquellas aplicaciones académicas e industriales futuras que se realicen en este contexto.}

\seplnkey{semantic textual similarity, annotation task, dataset, benchmark, cultural heritage, song lyrics}

\seplnabstract{In this paper, we present a new dataset and benchmark tailored to the task of semantic similarity in song lyrics. Our dataset, originally consisting of 2775 pairs of Spanish songs, was annotated in a collective annotation experiment by 63 native annotators. After collecting and refining the data to ensure a high degree of consensus and data integrity, we obtained 676 high-quality annotated pairs that were used to evaluate the performance of various state-of-the-art monolingual and multilingual language models. Consequently, we established baseline results that we hope will be useful to the community in all future academic and industrial applications conducted in this context.}

\firstpageno{1}

\begin{document}


\setlength\titlebox{17cm} 

\label{firstpage} 
\maketitle

%
    
\section{Introduction}


The success of music streaming services is mainly based on the tailor-made playlist they offer to users based on their listening habits and 54\%
of consumers say playlists are replacing albums in their listening habits\footnote{https://midiaresearch.com/blog/music-subscriber-market-shares-2022}. 
To build those lists and fit them to the user’s preferences, music streaming providers have developed recommender systems, which provide personalized suggestions based on the user´s behavior and some specific parameters related to the musicality of a song \cite{fell_natural_2020}, oftentimes overlooking the song lyrics. 

A song is comprised of two distinct components: the music and the lyrics. The lyrics consist of text, typically presented in poetic form, and 
describe the artist`s message and embody a unique combination of linguistic, artistic, and cultural elements, making them a special expression of writing that contains different features that could help us to improve 
song recommender systems. 
The lyrics thus provide a rich source of unstructured data and qualitative information that is not captured by metadata traditionally used in song recommendation such as beat, tempo, pitch, instrument, or mood (acoustic).

Due to their unique writing style, the song lyrics  present distinct challenges for semantic similarity modeling, 
a task that has traditionally been evaluated in the SemEVal tasks that were organized between 2012 and 2017 \cite{agirre_semeval-2012_2012,agirre_sem_2013,agirre_semeval-2014_2014,agirre_semeval-2015_2015}.
This evaluation framework operates under the assumption that a model that performs well for the general STS task is also likely to perform well for tasks that are tailored towards specific applications. 
However, there has been a significant discrepancy between the performance of models in STS and their performance in specific STS-based tasks such as MT Metrics (MTM) or Passage Retrieval (PR) \cite{abe_why_2022}. 
We argue existing STS-based evaluations\footnote{\url{http://nlpprogress.com/english/semantic_textual_similarity.html}} may not effectively capture their nuances and specificities, leading to suboptimal performance in lyric-related tasks such as recommendation, search, and cultural analysis, which hinders research and development efforts in this area. In addition, there is currently a gap in the availability of benchmark datasets that are specific to the domain of song lyrics \cite{chandrasekaran_evolution_2022}.

Therefore, a dataset of song lyrics annotated for similarity would not only facilitate the development of more accurate and applicable semantic similarity models but also enable a deeper understanding of the relationships between lyrics and music in this rich and diverse linguistic context. Since Spanish as a language contains a rich cultural and linguistic diversity and the Latin music industry represents a significant portion of the global music market, with a vast array of genres and styles, ranging from traditional folk music to contemporary pop, rock, or urban styles, the lyrics songs in Spanish offer unique challenges and opportunities for the exploration of semantic similarity in the domain of music.

In addition, the development of a dataset comprising Spanish song lyrics annotated for similarity tailored to the Spanish language and culture is crucial to advance the understanding and modeling of semantic similarity in the context of music and lyrics within the Spanish-speaking world.

In this paper, we introduce LyricSIM, a novel dataset for Spanish song lyric similarity designed to address the domain-specific characteristics of lyrics and facilitate the development of more accurate and applicable semantic similarity models for this domain. Our dataset comprises a diverse collection of paired song lyrics in Spanish, annotated with similarity scores based on various aspects such as theme, message, emotions, literal meaning, and cultural context.

On the other hand, we also have assessed the performance of various state-of-the-art models on our dataset for the semantic similarity task taking into account the unique features of song lyrics. 
The obtained results make up a new benchmark for the semantic similarity tasks based on the lyrics of songs. 
 
By developing a dataset specifically tailored to song lyric similarity and assessing the performance of SOTA models over this dataset, this paper aims to bridge the gap between general-purpose semantic similarity tasks and domain-specific applications, ultimately contributing to the advancement of NLP research in the context of music and lyrics research and analysis.

\section{Related Work}

To the best of our knowledge, no benchmark datasets currently exist for the study of semantic similarity between song lyrics. Although datasets containing song lyrics are available in different languages, they do not include semantic similarity annotations between song pairs, only collecting the lyrics. This information makes our dataset a very good fit to design models aimed at identifying plagiarism or detecting unlicensed versions of copyrighted songs, among others. 

The benchmark dataset we can consider most closely related to ours is the 4MuLA dataset introduced in \cite{10.1145/3428658.3431089}, which contains structured information that can be applied to several tasks. This dataset is obtained from a lyrics-focused platform and includes additional user-provided annotations. 
It includes 

Latin music genres that are underrepresented in other benchmark datasets. The dataset provides acoustic features, extracted tags, and lyrics in English, Portuguese, or Spanish, making it suitable for lyrics-, audio-, or multimodal-based genre classification, music and artist similarity, and popularity regression. Furthermore, the lyrics in the dataset can be used for cross- or multilingual text analysis, such as discourse analysis or measuring the differences between emotion transmitted by audio and lyrics. However, it lacks annotations about the semantic similarity between the song lyrics, which is what the benchmark dataset we propose provides.

Given that song lyrics often share similarities with poetry more than other literary genres, it stands to reason that poetry evaluation collections could be leveraged in the study of similarity between song lyrics, even when accounting for the differences between songs and poems.

In \cite{li2021CCPM}, a dataset was proposed in order to evaluate the semantic understanding of poetry models through poem matching. The objective was to advance research efforts focused on integrating deep semantics into the generation and comprehension system of Chinese classical poetry.
In \cite{haider-etal-2020-po} the annotation with experts leaded to an agreement of kappa = .70, resulting in a dataset for large scale analysis. The authors conducted first emotion classification experiments based on BERT, showing that identifying aesthetic emotions was challenging.

\section{Annotation Task}
Given that this was our first attempt at characterizing similarity in song lyrics, and that we wanted to obtain a broad vision of the problem, we decided to limit the number of annotators to three per pair and to favor diversity in the song lyrics. Through a crowdsourcing platform, participants were chosen from a pool of a total of 63 annotators who took part in the study. They were asked to rate the similarity between pairs of lyrics of Spanish songs using a six-point semantic differential scale ranging from 0, for completely different items, to 5 for outstandingly similar items (see details in Section \ref{sec:semantic-scale}). They were instructed to evaluate the similarity of pairs of song lyrics based on various criteria, such as the primary theme or context of the lyrics, the message conveyed, the emotions or feelings expressed, the literal meaning, the vocabulary employed, the relationship between the sender and receiver, the language style, and the sociocultural context of the song (see Annex 1 and 2 for a complete task description in Spanish and a translation into English, respectively). Participants were advised to follow their intuition if they had doubts or if the instructions were insufficient to provide a response. The aim of this task was to obtain a comprehensive dataset of similarity annotations for Spanish song lyrics, taking into account multiple dimensions of lyrical content.

\subsection{Dataset Description}
\begin{figure}[h]
  \centering
  \includegraphics[width=\linewidth]{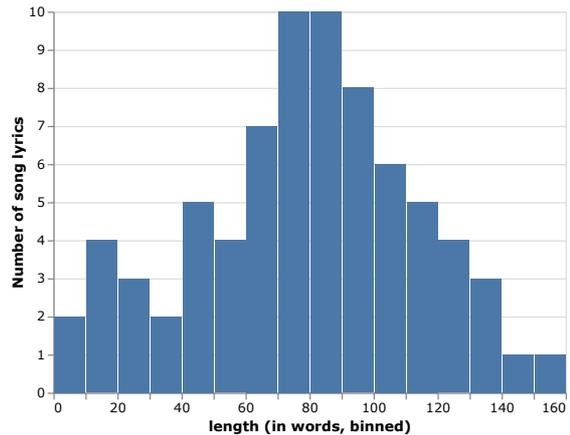}
  \caption{Histogram of song lyrics length in the dataset used to conduct the annotation task. The lengths were approximately normally distributed with a mean length of $77.61$ and a standard deviation of $34.88$.}
  \label{fig:lyrics_length_histogram}
\end{figure}

The dataset we prepared for the annotation task contains 75 song lyrics in Spanish that were selected for their diversity and popularity and for representing a wide range of music genres and themes. Also, we included song lyrics of varying lengths ($M=77.61, SD=34.88$, see histogram in Figure \ref{fig:lyrics_length_histogram}), to check whether this variable had any influence in the participants' perception of similarity. 

\begin{figure}[h]
  \centering
  \includegraphics[width=\linewidth]{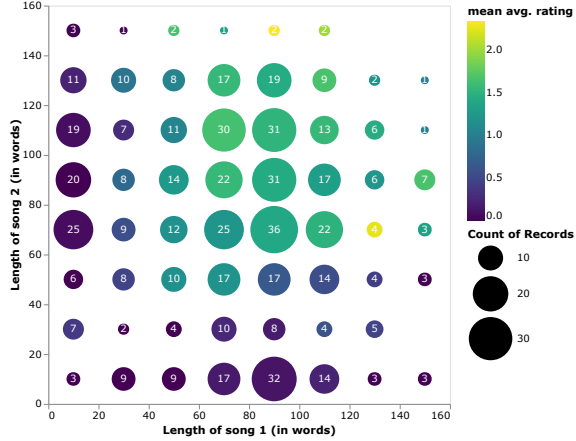}
  \caption{Scatterplot showing composition of pairs in the filtered dataset. The circles represent the size of combination cases (axes X and Y). The size of the circle depicts the number (in white) of particular cases matching lyrics of length $X$ to lyrics of length $Y$. The color scale represents the average of mean scores for items in that }
  \label{fig:song_vs_song_sim_heatmap}
\end{figure}

Although pairs were randomly assigned from the pool of 75 songs, we ensured that enough pair combinations with disparate lengths were paired together. A representation of these pairs can be seen in Figure \ref{fig:song_vs_song_sim_heatmap}.

\subsection{Semantic Differential Scale}
\label{sec:semantic-scale}
We employed a 6-point numeric scale in our annotation experiment, similar to the one that was conceived for the SemEval tasks \cite{agirre_semeval-2012_2012}. It is worth noting that, as in the original SemEval tasks, our scale devotes one level to total dissimilarity (level 0), and five other different grades to capture a subtler range of semantic similarity (levels 1-5) in an increasing order of intensity. However, we changed the wording of the categories to fit the broader context of similarity between song lyrics. A description of the full scale is provided below:

\begin{itemize}
    \item \textbf{Completely different (0)}: the lyrics are entirely dissimilar.
    \item \textbf{Barely any similarity (1)}: the lyrics share minor aspects without semantic importance, such as language style or sociocultural context.
    \item \textbf{Little similarity (2)}: there is no semantic similarity (lyrical situation, message, feelings), but the lyrics can be considered thematically (literal meaning) related.
    \item \textbf{Basic similarity (3)}: the lyrics resemble each other in message, feelings of the protagonist/singer, lyrical situation, or literal meaning.
    \item \textbf{Notable similarity / missing details (4)}: the lyrics share the same message and feelings but differ in lyrical situations and/or literal meaning.
    \item \textbf{Outstanding similarity (5)}: the lyrics share the same message, emotions, intentions, and lyrical situation, differing only in lexicon and genre. 
\end{itemize}

\begin{figure}[h]
  \centering
  \includegraphics[width=\linewidth]{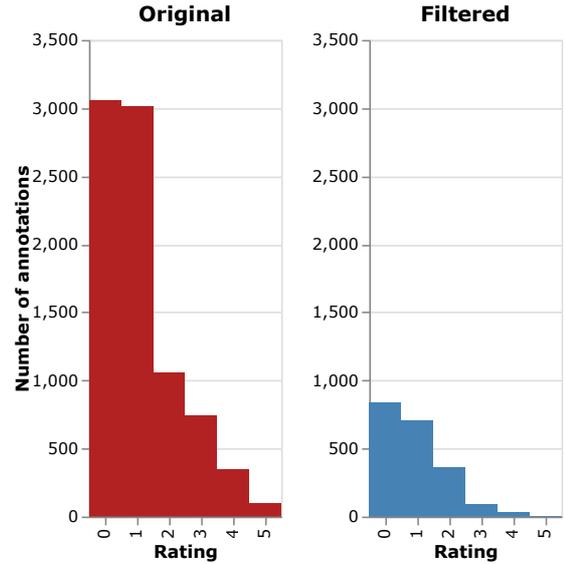}
  \caption{Comparison of the number of labels before (left, in blue) and after (right, in red) applying the filtering criteria. Although the difference in size between the two datasets is notable (approximately 75\% of the original annotations were discarded in the process), the filtered dataset contains only high-quality annotations.}
  \label{fig:dataset_filtering_comparison}
\end{figure}

Upon collection of the results, we obtained more 8,325 pair-wise similarity values corresponding to the comparison of 2775 pairs by three different participants each. An overview of the collected annotations can be seen to the left of Figure \ref{fig:dataset_filtering_comparison}.

\section{Data Refinement}
To maximize cost-effectiveness, we did not set any restrictions on the number of pairs that could be annotated by each participant, which resulted in an unbalanced distribution of annotation authoring. To mitigate this potential bias, we took the necessary steps to ensure that the resulting dataset included annotations that could be used as ground truth in future studies. This meant that we kept only those pairs in which a high degree of consensus (see \ref{subsec:consensus}) between the three annotators could be established.

 \begin{figure}[h!]
  \centering
  \includegraphics[width=\linewidth]{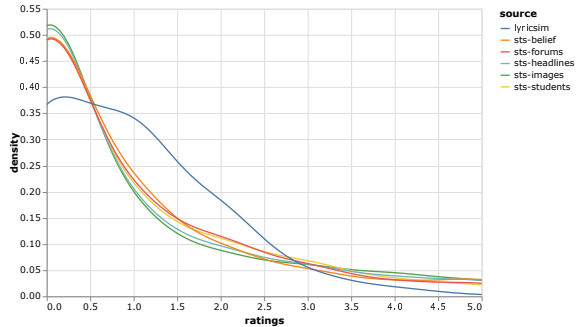}
  \caption{Comparison of rating distributions between the STS datasets and LyricSIM via kernel density estimation (KDE). In both cases, despite differences in the scale interpretation, the distributions are positively skewed, with STS datasets showing a higher bias towards the first point of the scale.}
  \label{fig:ratings_kernel_comparison}
\end{figure}

\subsection{Refinement}
In order to obtain high-quality annotations, we filtered the collected data using different criteria aimed at reducing rating variability. To model and filter the annotation data, as in the STS Core tasks, we departed from the assumption that annotations of similarity (scores from 1 to 5) are fundamentally different from those of dissimilarity (0). This effect can be seen in Figure \ref{fig:ratings_kernel_comparison}, which shows the kernel density estimations (KDEs) of rating distributions in the STS and LyricSIM datasets. The chart reveals that in all cases, ratings are biased towards the inferior ratings of the scale, specially in the STS data (around 50\% probability of seeing a rating of 0). Thus, we relied on the observation that the probability of seeing a dissimilar pair was hypothetically much higher, and consequently we made a clear distinction between dissimilar and similar items in an attempt to further characterize the tail of the distribution.

For dissimilarities, we take only those pairs in which all three annotators agreed that the pair was dissimilar (i.e., all annotators assigned a score of 0 to the pair), giving out a total of 837 pair-wise dissimilarities. In case of similarities, we aimed to capture pairs in which 2 out of 3 annotators agreed exactly on the same score. To avoid including doubtful cases, we chose to exclude pairs in which the third annotator assigned a very different score (i.e., 2 or more points apart from the mode), resulting in the selection of 676 high-quality similarity pairs or 24.36\% of the original dataset. In total, the refined dataset contains 2,028 high-quality annotations of pairwise similarity and dissimilarity judgments for a total of 75 distinct song lyrics. The 676 high-quality pairs are compared side by side to the unfiltered data in \ref{fig:dataset_filtering_comparison}, providing a visual estimation of the reduction in size per rating label in the refined dataset. This same data can be found in table format in Table \ref{tab:ratings}. 

\begin{table}[ht!]
\begin{center}
\begin{tabular}{c|c|c|c}
\hline
\textbf{dataset} & \textbf{rating} & \textbf{count} & \textbf{percent} \\
\hline
\multirow{6}{*}{Original} & 0 & 3058 & 36.73\% \\
                   & 1 & 3014 & 36.20\% \\
                   & 2 & 1058 & 12.71\% \\
                   & 3 &  746 &  8.96\% \\
                   & 4 &  347 &  4.17\% \\
                   & 5 &  102 &  1.23\% \\
\hline
\hline\rule{-2pt}{10pt}
\multirow{6}{*}{Filtered} & 0 & 837 & 41.27\% \\
                   & 1 & 705 & 34.76\% \\
                   & 2 & 360 & 17.75\% \\
                   & 3 &  88 &  4.34\% \\
                   & 4 &  34 &  1.68\% \\
                   & 5 &  4 &  0.20\% \\

\hline
\end{tabular}
\end{center}
\caption{\label{tab:ratings}Number of ratings in the original and filtered datasets.}
\end{table}

\subsection{Reliability}
\label{subsec:consensus}
To further assess the reliability of our dataset, we calculated inter-annotator agreement using Krippendorff's reliability alpha ($\alpha_{k}$), which gave a value of 0.90 (in the unrefined data, it was 0.27). Krippendorff's reliability alpha is a metric that generalizes other metrics that are responsible for quantifying the reliability between annotators (inter-rater reliability). It can be used for both ordinal and nominal annotations, as well as with any number of annotators. K-alpha yields a value between 0 and 1, where 1 represents full agreement. However, there are different criteria regarding when to consider that there is enough agreement between annotators. According to the general consensus, a common threshold is to consider that there is enough agreement when K-alpha is greater than 0.8. Using this metric, the resulting inter-annotator agreement of the high-quality annotations dataset was found to be substantial, with a coefficient alpha of 0.90, indicating that the dataset is reliable and thus can be used for future research in the field. Table \ref{tab:krippendorff-sts} shows a comparison between STS gold standard datasets used in the SemEval 2014 STS Core task \cite{agirre_semeval-2014_2014} and ours, including dataset size (in sentence pairs), number of annotators per pair and computed Krippendorff's alpha-reliability scores \cite{krippendorff_reliability_2004}, when possible. Unfortunately, we could not find the results of the STS Spanish tasks disaggregated by annotator. Thus, we derived our results from the English tasks data (available at \url{http://ixa2.si.ehu.es/stswiki/images/2/21/STS2015-en-rawdata-scripts.zip}). 

\begin{table} [h]
\begin{center}
\begin{tabular} {l|c|c|c}
  \hline
  {\bf Dataset} & {\bf Size} &  {\bf Avg.Length} & {\bf K-\textalpha} \\
  \hline
  STS-images & 1500 & $9.6\pm3.04$ & 0.82\\
  STS-students & 1500 & $10.44\pm3.34$ & 0.72\\
  STS-headlines & 2000 & $7.5\pm2.24$ & 0.79\\
  STS-belief & 2000 & $13.01\pm6.83$ & 0.63\\
  STS-forums & 1500 & $15.04\pm3.28$ & 0.66\\
  \hline
  Avg. STS & & & 0.72\\
  \hline
  LyricSIM & 676 & $77.61\pm34.88$ & 0.90\\
  \hline
\end{tabular}
\end{center}
\caption{\label{tab:krippendorff-sts}Comparison between STS datasets and LyricSIM displaying dataset size (in pairs), average sentence/utterance length and calculated Krippendorff alpha (K-alpha). Text sizes in LyricSIM are considerably larger than their STS counterparts and K-alpha varies greatly in the STS datasets, ranging from 0.63 (min) to 0.82 (max).}
\end{table}

\section{Evaluation}
In this section, we use the refined dataset introduced in the previous section to analyze how SOTA language models perform in the similarity detection task. We have followed a similar 85-5-10 split as the one used in other studies \cite{gutierrez-fandino_maria_2022,agerri_lessons_2023}, resulting in 638 song pairs for the train set, 38 song pairs for the development set and 68 song pairs for the test set. To ensure balanced representation of each class, we used stratified sampling during the splitting process. The metrics used for model assessment were Spearman's Rank Correlation and Pearson's Correlation \cite{cer_semeval-2017_2017,devlin_bert_2019}. For the purpose of creating a reference point for our dataset's behavior when subjected to traditional, less complex models, a Support Vector Machine (SVM) model was also trained. Further, to obtain a more holistic measure of the models' performance, a combined score was also calculated, computed by taking the arithmetic mean of both metrics \cite{gutierrez-fandino_maria_2022} \cite{agerri_lessons_2023}.

\subsection{Language Models}
\label{sec:language-models}
We present a brief overview of the models that were selected for evaluation and discuss their respective training and architectural characteristics. By exploring these models in greater depth, we can better understand the nuances of their performance on the refined dataset, ultimately allowing us to make informed decisions regarding their suitability for similarity detection tasks in the Spanish language.


The language model architectures we have selected are BERTIN, RoBERTa-base-bne (MarIA base), RoBERTa-large-bne (MarIA large), Sentence Transformer, ALBERTI, DeBERTa, XML-RoBERTa base and XML-RoBERTa large. There are several reasons that led us to choose these five models for our study. First, we aimed to include a significant representation of both monolingual Spanish models, as our dataset comprises Spanish song lyrics, and multilingual models, which have demonstrated superior performance over monolingual models in Spanish tasks. Additionally, we selected both the base and large versions of these models to analyze the resulting metrics after training.

Our selection was influenced by an article that evaluated Spanish Language Models\cite{agerri_lessons_2023}. The models we chose exhibited high performance in terms of their STS official combined scores, as shown in Table 2 of the referenced article. BERTIN and MarIA have become prominent monolingual models for the Spanish language, as they have been trained on extensive Spanish datasets. BERTIN was trained on the Spanish portion of mC4, which contains approximately 416 million documents and 235 billion words in 1TB of uncompressed data, along with other datasets such as Wikipedia, OpenSubtitles, and Europarl. MarIA was trained on a 570GB corpus of clean and deduplicated texts extracted from the Spanish Web Archive, built by the National Library of Spain between 2009 and 2019.

Sentence Transformer was chosen due to its core focus on STS tasks during training, which is highly relevant to our study. ALBERTI was selected because it is domain-adapted for poetry, a semantic context closely related to song lyrics. The reasons for choosing DeBERTa, XML-RoBERTa base, and XML-RoBERTa large models are manifold. DeBERTa is selected due to its superior training scores compared to RoBERTa and BERT, incorporating cutting-edge techniques such as Disentangled Attention and Enhanced Mask Decoder, which enhance its performance. XML-RoBERTa base and XML-RoBERTa large models are chosen for their extensive parameter counts, with the base version comprising 270M parameters and the large version containing 550M parameters. Additionally, these models are multilingual, trained on 100 languages, making them highly versatile and capable of handling diverse language tasks.

In the following, we provide an overview of these models representing both monolingual and multilingual approaches. All these models were trained on various corpora with distinct architectures and training parameters which we discuss in the following section. 

\begin{itemize}
    \item BERTIN \cite{rosa_bertin_2022} employs a novel technique called "perplexity sampling" for pre-training Spanish language models. This method reduces the amount of data and training steps needed compared to traditional approaches, while still achieving competitive results. BERTIN utilizes a two-step training process with different sequence lengths and batch sizes.
    \item The MarIA \cite{gutierrez-fandino_maria_2022}, RoBERTa-base-bne and RoBERTa-large-bne models \cite{liu_roberta_2019}, are state-of-the-art NLP models for the Spanish language. They have been trained on a massive corpus of Spanish text data derived from the National Library of Spain's selective crawls. The MarIA models employ a single-epoch training approach with no dropout, focusing on tasks such as sentiment analysis, part-of-speech tagging, and named entity recognition.
    \item Sentence-Transformer \cite{reimers_sentence-bert_2019} is a modification of pre-trained BERT and RoBERTa networks that produce semantically meaningful sentence embeddings. It uses a siamese and triplet network structure, reducing computational overhead while maintaining accuracy.
    \item ALBERTI\footnote{\url{https://huggingface.co/flax-community/alberti-bert-base-multilingual-cased}} is a BERT-based multilingual model trained on poetry datasets, including Spanish resources. Although no publication detailing the corpus and training methodology exists, the model leverages domain adaptation to capture patterns and features specific to poetry.
    \item DeBERTa \cite{he_deberta_2021}, Decoding-enhanced BERT with Disentangled Attention, refines BERT and RoBERTa through disentangled attention and superior masked decoding. The model's virtual adversarial training enhances generalization, improving efficiency in pre-training and performance in NLU and NLG tasks. An expansive DeBERTa version surpasses human performance on the SuperGLUE benchmark, a notable achievement in macro-average score.
    \item XML-RoBERTa \cite{conneau_unsupervised_2020} enhances cross-lingual understanding (XLU) through extensive study of unsupervised cross-lingual representations. Introducing XLM-R, a transformer-based multilingual model pre-trained on text in 100 languages, achieving state-of-the-art results in cross-lingual classification, sequence labeling, and question answering. The authors delve into key factors and trade-offs between positive transfer, capacity dilution, and performance across languages. They demonstrate that a single large model can effectively encompass all languages without sacrificing per-language performance.

\end{itemize}

\subsection{Training Parameters Details}

In this section, we delve deeper into the training parameters employed by the monolingual and multilingual models, highlighting the similarities and differences that may contribute to their respective performance in the similarity detection task.

We want to compile the training parameters in a list format, which we have obtained after reading the papers of each of these models. These parameters serve as a starting point for replicating and investigating the results obtained by the research and development teams who worked on each of these models. The training parameters for the selected models reveal substantial differences in the pre-training procedures employed by the monolingual and multilingual models. For example, BERTIN follows a two-step training process with varying sequence lengths and batch sizes, while the MarIA models employ a single-epoch training approach with no dropout. These variations in training procedures and corpora used for pre-training contribute to the distinct performance characteristics of each model. As researchers continue to develop and evaluate Spanish language models, it is crucial to assess how these differences in training parameters impact the models' effectiveness in downstream tasks. The selected Spanish language models offer valuable insights into the development and application of transformer-based language models for the Spanish language.

\subsection{Fine-tuning}
The fine-tuning process was carried out following the same practices found in the aforementioned studies. We used the same scripts as the MarIA team\footnote{https://github.com/PlanTL-GOB-ES/lm-spanish} \cite{gutierrez-fandino_maria_2022}, which are based on HuggingFace Transformers library \cite{wolf_transformers_2020}, with minor modifications to adapt them to our dataset structure. To maintain consistency across the models, we initialized each one of them with a random head and employed a fixed seed for reproducibility. We conducted a grid search over the following search space:
\begin{itemize}
    \item Weight decay: 0.1, 0.01
    \item Learning rate: 1e-5, 2e-5, 3e-5, 5e-5
    \item Batch size: 8, 16, 32
\end{itemize}

Due to memory constraints, especially for larger models, in cases where the batch size exceeded capacity, gradient accumulation was used achieving the same effective batch size; the rest of the hyperparameters remain the same as the HuggingFace defaults. The maximum sequence length was 512 tokens for all models, chosen to accommodate all sentence pairs in the dataset. To prevent overfitting, we trained each model for a maximum of 5 epochs with Adam optimizer and a linear decaying learning rate, selecting the checkpoint with the highest score according to development set. Finally, we perform evaluation on the test set for each model, with the best checkpoint from the previous step. The optimal configuration of hyperparameters per model is shown in Table \ref{tab:best_config_scores}.

\subsection{Results}
The results are shown in Table \ref{tab:test_scores} for each model, fine-tuned as described. Among the eight models investigated, MarIA large demonstrated the highest performance in terms of the combined score metric, while MarIA's base counterpart ranked fourth; mDeBERTa model closely followed MarIA large, exhibiting competitive results. The XLM-RoBERTa large and base models are fifth and sixth respectively, showing again a larger model performing better than its base variant, while Sentence Transformer based on the same architecture performed better than both, achieving third place. ALBERTI, the only BERT based model of this study, was only slightly behind some of the other models, placing seventh. BERTIN, although not dramatically behind the SVM, placed last. These findings suggest that the MarIA large model is the most effective in capturing semantic textual similarity on songs, with mDeBERTa as a strong contender.

The results for the evaluation on the development set, used as criteria for selecting the best checkpoint from the fine-tuning process and hyperparameter selection, can be found in Table \ref{tab:dev_scores} of the appendix. A comparative analysis of the development and test set results may provide further insights into the generalization capabilities of the models and the effectiveness of the fine-tuning process.
\begin{table*}[ht!]
\centering
\begin{tabular}{lcccc}
\toprule
    \textbf{model name} & \textbf{STS combined} & \textbf{combined} & \textbf{spearmanr} & \textbf{pearson} \\
\midrule
    BERTIN &                79.45  &          85.72 &        86.32 &      85.12 \\
    MarIA large &            84.11 & \textbf{90.02} & \textbf{89.94} & \textbf{90.11} \\
    MarIA base  &   \textbf{85.33} &          86.75 &        89.02 &      84.5 \\
    XLM-RoBERTa base &       83.47 &          86.45 &        88.19 &      84.72 \\
    XLM-RoBERTa large  &    84.04  &          86.74 &        88.15 &      85.33 \\
    Sentence Transformer XLM-R  & -  &        88.91 &        89.55 &      88.28 \\
    mDeBERTa3  &             83.61 &          89.15 &        89.07 &      89.23 \\
    ALBERTI  &                     - &        86.35 &        88.29 &      84.42 \\
    SVM (RBF)  &                     - &        86.18 &        86.57 &      85.78 \\
\bottomrule
\end{tabular}
\caption{\label{tab:test_scores}Test combined scores for all the models considered (best in bold). STS dataset results from other studies \cite{agerri_lessons_2023} have been added for illustration purposes.}
\end{table*}

\section{Discussion}
In this section, we will delve into the results obtained from the data refining process and the fine-tuning of the baseline models, highlighting the insights and implications of these outcomes for our understanding of semantic similarity in Spanish song lyrics. 

\subsection{Evaluation Results}
It is noteworthy that the highest performing model, MarIA large, is a Spanish-specific model. This observation suggests that models trained on a particular language may have an advantage in capturing the subtleties of that language for this particular task, although multilingual models can still achieve competitive performance. Further studies are recommended to provide a more comprehensive understanding. Furthermore, our results validate that larger models (e.g. Maria large) lead to improved performance compared to their base counterparts. This result emphasizes the need to factor in model size and computational demands when choosing a model for practical applications. 

Another interesting observation is the difference in performance between the BERT-based model, ALBERTI, and the other RoBERTa and XLM-RoBERTa-based models. Surprisingly, even though this model was trained on poems, which exhibit a structure more closely related to songs than the data used to train the other models, performs relatively worse. A possible avenue for exploration would be the tokenization process used by this model, as it does not take into account line-break characters (\textbackslash n) commonly used to delimit verses and stanzas, which in turn could impact the model's ability to learn the structure of a song. This also accentuates the importance of considering the structure and formatting of the input data when selecting a model for a specific task. 

Similarly, the advantage in performance of Sentence Transformer, which is based on the XLM-RoBERTa architecture, in comparison to the other XLM-RoBERTa models may be partially attributed to the supplementary fine-tuning process applied to sentence pairs for semantic textual similarity. We argue this fine-tuning might have the potential to refine the model's ability to capture semantic relationships more effectively, which would explain this observation in the data.

The results also revealed that the performance of the SVM model, a comparatively simpler machine learning architecture, was not significantly different from some of the more complex transformer-based models (e.g. BERTIN). This could potentially indicate that the dataset size might not be large enough or sufficiently complex to highlight the strengths of transformer models.

These findings have practical implications for the development of real-world applications related to the task at hand: the performance of the MarIA large and mDeBERTa models suggests that they may be well-suited for tasks such as song recommendation, lyric analysis, or music information retrieval.

\subsection{Similarity vs Length}
\begin{figure}[h!]
  \centering
  \includegraphics[width=\linewidth]{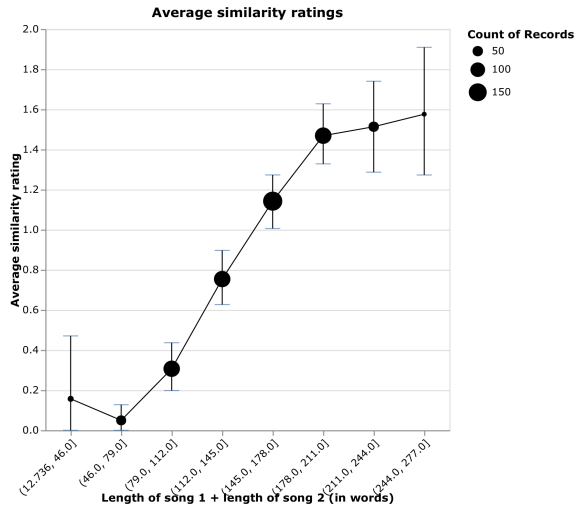}
  \caption{X axis: word sum of pairs in the annotation dataset. Y axis: average similarity rating with 95\% CI error bars. Despite disparity in sample size across lengths, in our experiment, the perceived similarity in song lengths increased with the lyrics length, indicating a possible interaction effect.}
  \label{fig:sim_ratings_by_songs_length}
\end{figure}

In addition to the findings discussed earlier, an interesting observation that we could derive from the annotation data was that similarity scores increased with the total length of the lyrics in a pair (see Figure \ref{fig:sim_ratings_by_songs_length}), suggesting that longer lyrics may contain more opportunities for shared vocabulary and thematic elements  to contribute to overall semantic similarity. Although more research is needed in this area, current evidence points to the potential utility of considering vocabulary overlap as an important factor in determining semantic similarity between pieces of text \cite{abe_why_2022}. As we discuss in the next section, there is a currently a lack of examples of annotated pairs exhibiting varying degrees of similarity (low, medium or high), which we aim to resolve in future studies.

\section{Conclusion and Future Work}
In this study, we have laid the groundwork for assessing semantic similarity in the context of Spanish song lyrics by presenting a dataset and baseline results from pre-trained SOTA models. Our results provide insights into the performance of these models, revealing potential strengths and weaknesses as well as opportunities for future research. Beyond that, we have provided a reflection on the data collection process typical of the similarity annotation task, and a detailed characterization of the filtering process that we followed to increase the quality of the annotation data. Finally, we provide all the code and data necessary to reproduce our research at the repository located at 
\url{https://github.com/linhd-postdata/lyricsim}. 

To further advance this research area, during the course of our research we identified several avenues for future work that can be pursued. One possible direction is expanding the dataset to make it multilingual by incorporating song lyrics from different languages and cultures. Furthermore, it is essential to investigate the appropriateness of the psychometric scales typically used in similarity annotation studies, as different scales or measurement techniques could prove more effective in capturing and quantifying semantic similarity in song lyrics.

Addressing the issue of imbalanced datasets, it is necessary to collect more data on similarity pairs, composing them in a manner that allows for an approximately uniform distribution of similarity ratings. This can be achieved, for example, by precalculating similarities using one of the models discussed in this paper. Building on the knowledge that language models can partially capture semantic similarity between song lyrics, researchers can design more targeted experiments to gain a deeper understanding of the underlying factors that contribute to similarities and differences in lyrical content, such as the relationship between song length and similarity detection.

Moreover, conducting more user studies is crucial to assess the reliability and validity of the models' inferences. These studies should encompass not only the annotation task and the training and evaluation processes of the models, but also a user-driven evaluation of the models' performance in real-world applications, a practice that is already attracting the interest of the NLP community \cite{schuff_how_2023}. Lastly, continued analysis of the experimental results obtained in this study will lead to a better understanding of the strengths and limitations of the models used for capturing semantic similarity in song lyrics, continuing with recent research in the field \cite{zhang_optimal_2021}.

To summarize, this research has made significant progress in comprehending semantic similarity within Spanish song lyrics, laying the groundwork for further investigation. By addressing potential areas of future work, as mentioned above, our aim is to develop a more all-encompassing understanding of the factors that influence similarities and differences in lyrical content across diverse languages and cultures. Ultimately, these advancements will enhance models and applications for capturing and measuring semantic similarity within song lyrics not only for the Spanish-speaking community but also on a global scale. As we persist with refining our methods and exploring new approaches, we remain devoted to promoting natural language processing advancement while appreciating human linguistic diversity as well as musical expression.

\section*{Acknowledgments}
\begin{itemize}
   \item This research has been carried out in the framework of the Grant LyrAIcs Grant agreement ID: 964009 funded by ERC-POC-LS
   \item This research has been carried out in the framework of the Grant CLS INFRA reference 101004984 funded by H2020-INFRAIA-2020-1. 
   \item This research has received funding from the project \textit{ISL: Intelligent Systems for Learning} (GID2016-39) in the call PID 22/23 of the Vice-rectorate for Educational Innovation of the University for National Distance Education of Spain (UNED).
   \item Alejandro Benito-Santos acknowledges support from the postdoctoral grant "Margarita Salas", awarded by the Spanish Ministry of Universities. 
\end{itemize}

\bibliography{references}
\bibliographystyle{fullname_esp}

\appendix
\section{Annex 1: Original Task Description in Spanish} 
\label{sec:annex-1}

Instrucciones: Para esta tarea, comparará las letras de 2 canciones y evalúa su similitud teniendo en cuenta las siguientes reglas de interpretación de las letras:

\begin{itemize}
    \item Tema principal / contexto de la letra (por ejemplo: guerra, engaño, conocer a una nueva persona)
    \item Mensaje: opinión destacable del tema de la letra (por ejemplo: "La vida es muy corta", "Disfruta tu vida", "Uno puede amar solo una vez")
    \item Sentimientos / emociones del tema de la letra (decepción, tristeza, esperanza,enamoramiento)
    \item Significado literal de la letra /del vocabulario / ambiente (cosmos, navegación, prado, vida en la gran ciudad, etc.)
    \item Relación entre el emisario y el receptor de la letra (el hombre le canta a una niña, etc.)
    \item Estilo de lenguaje (jerga de adolescentes)
    \item Conocimiento sobre el contexto social cultural de la canción (biografía del autor/género musical, edad de la creación, etc.)
\end{itemize}

Teniendo en cuenta estas consideraciones puntúa la similitud ente los pares de letras de canción entre 0 y 5 siendo:

\begin{enumerate}
    \setcounter{enumi}{-1}
    \item \textbf{Completamente diferentes}: Las letras son totalmente diferentes.
    \item \textbf{Apenas existe similitud}: se parecen en cualquiera de los cualquiera de los aspectos sin importancia semántica como el estilo del lenguaje, o el contexto social/cultural del contenido de las letras.
    \item \textbf{Poca similitud}: no hay similitud semántica (situación lírica, mensaje, sentimientos) pero las letras se pueden considerar temáticamente (significado literal) relacionadas entre sí.
    \item \textbf{Similitud básica}: solo se parecen en el mensaje, sentimientos del protagonista/cantante, situación lírica o significado literal.
    \item \textbf{Similitud notable/faltan detalles}: el par de letras comparten el mismo mensaje y sentimientos del sujeto, pero difieren en situaciones líricas y/o significado literal.
    \item \textbf{Similitud Sobresaliente}: El par de letras tienen el mismo mensaje, comparten la misma situación lírica, emociones e intenciones. La diferencia solo puede estar en el léxico y en los géneros.
\end{enumerate}

CONSIDERACION FINAL: Si tiene dudas y las instrucciones no son suficientes para responder, siga su intuición. 

\section{Annex 2: Task Description Translated into English}
\begin{itemize}
\item Main theme/context of the lyrics (e.g., war, deceit, meeting a new person)
\item Message: notable opinion on the subject of the lyrics (e.g., "Life is very short," "Enjoy your life," "One can love only once")
\item Feelings/emotions of the subject of the lyrics (disappointment, sadness, hope, infatuation)
\item Literal meaning of the lyrics/vocabulary/setting (cosmos, sailing, meadow, life in the big city, etc.)
\item Relationship between the sender and the receiver of the lyrics (man singing to a girl, etc.)
\item Language style (teenage slang)
\item Knowledge about the sociocultural context of the song (author's biography/music genre, age of creation, etc.)
\end{itemize}

Considering these factors, rate the similarity between the pairs of song lyrics from 0 to 5 as follows:

\begin{enumerate}
    \setcounter{enumi}{-1}
    \item \textbf{Completely different}: the lyrics are entirely dissimilar.
    \item \textbf{Barely any similarity}: the lyrics share minor aspects without semantic importance, such as language style or sociocultural context.
    \item \textbf{Little similarity}: there is no semantic similarity (lyrical situation, message, feelings), but the lyrics can be considered thematically (literal meaning) related.
    \item \textbf{Basic similarity}: the lyrics resemble each other in message, feelings of the protagonist/singer, lyrical situation, or literal meaning.
    \item \textbf{Notable similarity / missing details}: the lyrics share the same message and feelings but differ in lyrical situations and/or literal meaning.
    \item \textbf{Outstanding similarity}: the lyrics share the same message, emotions, intentions, and lyrical situation, differing only in lexicon and genre. 
\end{enumerate}

FINAL CONSIDERATION: If you have doubts and the instructions are not sufficient to provide an answer, follow your intuition.

\section{Annex 3: Summary of training parameters}
\label{sec:annex_3}
A summary of the training parameters for each of the selected models is provided below:

\begin{itemize}
    \item BERTIN
    \begin{itemize}
        \item Total steps: 250k
        \item Step 1:
        \begin{itemize}
            \item Steps: 230k
            \item Sequence length: 128
            \item Batch size: 2048
        \end{itemize}
        \item Step 2:
    \begin{itemize}
    \item Steps: 20k
    \item Sequence length: 512
    \item Batch size: 384
    \end{itemize}
    \item Epochs: Not specified
    \end{itemize}

\item RoBERTa-base-bne (MarIA base)
\begin{itemize}
\item Total steps: Not specified
\item Sequence length: 512
\item Batch size: 2048
\item Epochs: 1
\item Adam optimization with $\beta_1 = 0.9$, $\beta_2 = 0.98$, $\epsilon = 1e^{-6}$, and L2 weight decay of 0.01
\end{itemize}

\item RoBERTa-large-bne (MarIA large)
\begin{itemize}
\item Total steps: Not specified
\item Sequence length: 512
\item Batch size: 2048
\item Epochs: 1
\item Adam optimization with $\beta_1 = 0.9$, $\beta_2 = 0.98$, $\epsilon = 1e^{-6}$, and L2 weight decay of 0.01
\end{itemize}

\item Sentence Transformer
\begin{itemize}
\item Total steps: 1.5M
\item Sequence length: 512
\item Batch size: 8192
\item Epochs: Not specified
\item Adam optimization with $\beta_1 = 0.9$, $\beta_2 = 0.999$, $\epsilon = 1e^{-4}$, and L2 weight decay of 0.01
\end{itemize}

\item ALBERTI
\begin{itemize}
\item Total steps: Not specified
\item Sequence length: Not specified
\item Batch size: Not specified
\item Epochs: Not specified
\end{itemize}

\item DeBERTa
\begin{itemize}
\item Total steps: 1M
\item Sequence length: 768
\item Batch size: 64
\item Epochs: 10
\item Adam optimization with $\beta_1 = 0.9$, $\beta_2 = 0.999$, $\epsilon = 1e^{-6}$, and L2 weight decay of 0.01
\end{itemize}

\item XML-RoBERTa
\begin{itemize}
\item Total steps: Not specified
\item Sequence length: Not specified
\item Batch size: 8192
\item Epochs: Not specified
\end{itemize}

\end{itemize}

\section{Annex 4: Table of test combined scores}
\begin{table*}[]
\centering
\begin{tabular}{lcccc}
\toprule
    \textbf{model name} &  \textbf{combined} &  \textbf{spearmanr} &  \textbf{pearson} \\
\midrule
       BERTIN &                87.61	& 88.19	& 87.04\\
       MarIA large  &          89.12	& 89.97	 & 88.26 \\
       MarIA base &            87.65	& 89.46 &	85.84 \\
       XLM-RoBERTa base &            \textbf{90.63}	& \textbf{92.23}	& \textbf{89.03} \\
       XLM-RoBERTa large &         89.04	 & 89.34	 & 88.74 \\
       Sentence Transformer XLM-R &       88.31 &	89.46 &	87.17 \\
       mDeBERTa3 &            88.14	& 88.59 &	87.68 \\
       ALBERTI &            87.65	& 88.92 &	86.38\\
\bottomrule
\end{tabular}
\caption{\label{tab:dev_scores}Development set combined scores for all the models considered.}
\end{table*}

\begin{table*}[]
\centering
\begin{tabular}{lccccc}
\toprule
    \textbf{model name} &  \textbf{Batch size} &  \textbf{Weight decay} &  \textbf{Learning rate} & \textbf{Eval} & \textbf{Test} \\
\midrule
       BERTIN &                32 & 0.01 &  0.00005 &  87.61 &  85.72\\
       MarIA large  &          32 & 0.1 &   0.00003 &  89.12 &   \textbf{90.02}\\
       MarIA base &             8 & 0.01 &  0.00003 &  87.65 &      86.75\\
       XLM-R base &             8 & 0.1 &   0.00001 &  \textbf{90.63} &  86.45\\
       XLM-R large &           16 & 0.01 &   0.00005 &  89.04 &  86.74\\
       ST XLM-R &              8 & 0.01 &  0.00005 &  88.31 &  88.91\\
       mDeBERTa3 &             16 & 0.01 &   0.00005 &  88.14 & 89.15\\
       ALBERTI &               16 & 0.1 &   0.00003 &  87.65 & 86.35\\
\bottomrule
\end{tabular}
\caption{\label{tab:best_config_scores}Best configuration for each model with combined test and evaluation scores.}
\end{table*}

\end{document}